\documentclass[sn-basic]{sn-jnl} 
\usepackage[utf8]{inputenc}
\usepackage{array,tikz}
\usepackage{graphicx}%
\usepackage{multirow}%
\usepackage{amsmath,amssymb,amsfonts}%
\usepackage{amsthm}%
\usepackage{mathrsfs}%
\usepackage[title]{appendix}%
\usepackage{xcolor}%
\usepackage{textcomp}%
\usepackage{manyfoot}%
\usepackage{booktabs}%
\usepackage{algorithm}%
\usepackage{algorithmicx}%
\usepackage{algpseudocode}%
\usepackage{listings}%
\begin{document}

\title{Explainable Generative AI (GenXAI): A Survey, Conceptualization, and Research Agenda} 
\author{\fnm{Johannes} \sur{Schneider}}\email{johannes.schneider@uni.li}
\affil{\orgname{University of Liechtenstein}, \city{Vaduz}, \country{Liechtenstein}}

\abstract{
Generative AI (GenAI) marked a shift from AI being able to “recognize” to AI being able to “generate” solutions for a wide variety of tasks. As the generated solutions and applications become increasingly more complex and multi-faceted, novel needs, objectives, and possibilities have emerged for explainability (XAI). In this work, we elaborate on why XAI has gained importance with the rise of GenAI and its challenges for explainability research. We also unveil novel and emerging desiderata that explanations should fulfill, covering, for instance, verifiability, interactivity, security, and cost aspects. 
To this end, we focus on surveying existing works. Furthermore, we provide a taxonomy of relevant dimensions that allows us to better characterize existing XAI mechanisms and methods for GenAI.
We discuss different avenues to ensure XAI, from training data to prompting. Our paper offers a short but concise technical background of GenAI for non-technical readers, focusing on text and images to better understand novel or adapted XAI techniques for GenAI. However, due to the vast array of works on GenAI, we decided to forego detailed aspects of XAI related to evaluation and usage of explanations. 
As such, the manuscript interests both technically oriented people and other disciplines, such as social scientists and information systems researchers. Our research roadmap provides more than ten directions for future investigation.
}
\maketitle

\section{Introduction} \label{sec:intro}
Generative AI (GenAI) has shown remarkable capabilities that have shaken up the world on a broad basis -- ranging from regulators~\citep{EU2023AIACT}, educators~\citep{bai23}, programmers~\citep{Sob23} onto medical staff~\citep{thir23}. 
For businesses~\citep{port23g}, GenAI has the potential to unlock trillions of dollars annually~\citep{Mc23Ge}. At the same time, it is said to threaten mankind~\citep{Guar23}. These opposing views are a key drive for understanding and explaining GenAI.
Generative AI constitutes the next level of AI driven by foundation models~\citep{schn24f}, where AI can create text, images, audio, 3D solutions, and videos~\citep{goza23,cao23} controllable by humans, specifically via textual prompts~\citep{Whi23} -- see also Table~\ref{tab:aisamp} for examples of public GenAI systems.
This is a major step forward from AI that was primarily able to ``recognize'' to AI that ``generates''. GenAI has shown unprecedented capabilities like passing university-level exams~\citep{cho21c,kat24}. It also achieves remarkable results even in areas deemed non-suitable for machines, such as creativity~\citep{Chen2023Hu}. It is accessible to everyone, as witnessed by commercial systems like ChatGPT~\citep{achi23gpt} and Dall-E~\citep{bet23,ram22h}. Early generative AI methods, such as Generative Adversarial networks (GANs), can also generate artifacts but are typically more difficult to control than modern models such as transformers and diffusion architectures. 

Explainable AI for GenAI (GenXAI) techniques produce explanations that help comprehend AI, for example, outputs for individual inputs or the model as a whole. Traditionally, explanations have served many purposes due to multiple needs; for instance, they can increase trust and support the debugging of models~\citep{mesk22}. The need for understanding AI is even larger than in pre-GenAI eras. For example, explanations can support the verifiability of generated content and, thereby, contribute to combatting one of the major problems of GenAI: hallucinations (as argued in Section~\ref{sec:und}). 
Unfortunately, Explainable AI (even for pre-GenAI models) is characterized by several open problems despite many attempts to design solutions to address them over the last few years~\citep{longo24,mesk22}. 
For example, a recent comparison~\citep{silv23} among methods on the impact of XAI on human-agent interaction found that the difference in scores between the best (counterfactuals) and worst method (using simply probability scores) was only 20\%, hinting that complex existing methods yield limited benefits over more complex ones. Thus, XAI techniques are still far from being optimal. 
Other works have even openly called the ``status quo in interpretability research largely unproductive''~\citep{rauk23}. As such, there is much to be done, and it is essential to understand current efforts to learn and improve on them -- especially to mitigate high risks~\citep{Guar23} while leveraging opportunities~\citep{schn24f}. 

\begin{figure}[h]
\centering
\includegraphics[width=0.7\textwidth]{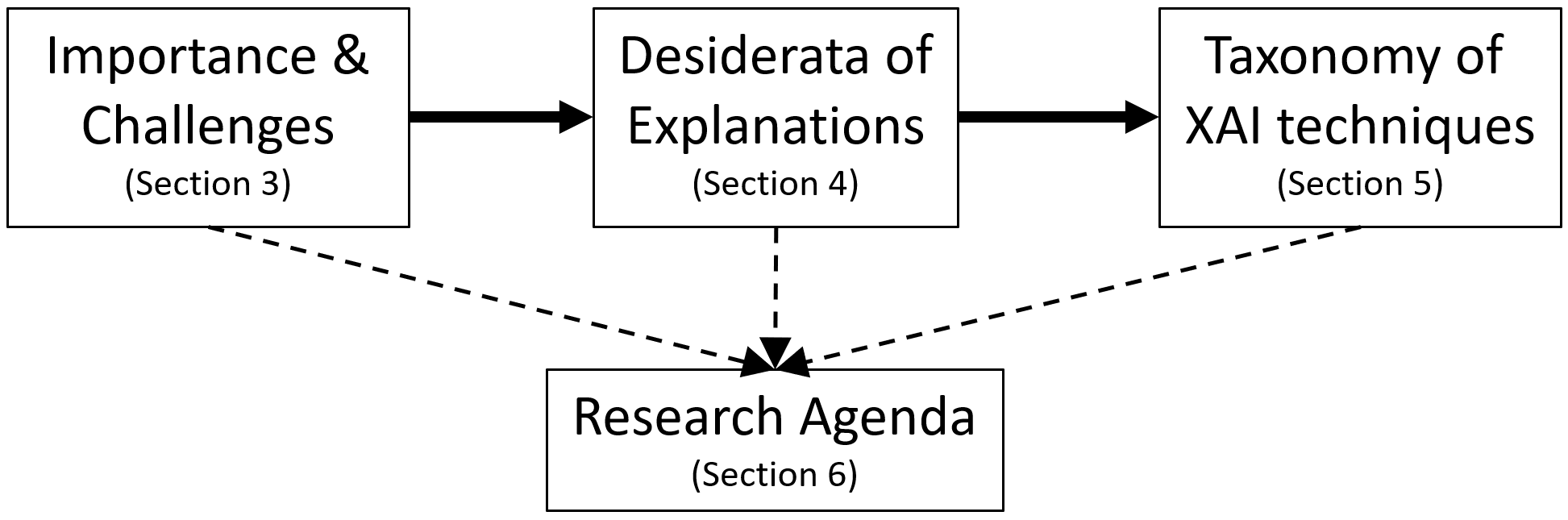}
\caption{Article outline. Following important factors and challenges of XAI for GenAI, desiderata and, in turn, a taxonomy is derived. All three together inform our research agenda.}\label{fig:out}
\end{figure}

This research manuscript is a genuine attempt to progress in this direction. Our goal is not to (only) list and structure existing XAI techniques, as in the current stage of the field, more basic questions need to be addressed, such as identifying key challenges and desiderata for GenXAI. To this end, we, therefore, opted for a more narrative review methodology~\citep{king05und} accompanied by a taxonomy development approach from the field of information systems~\citep{nick13}.\footnote{Details on the methodology are given in the Appendix (Section \ref{sec:resMeth}).}

There are several surveys on XAI focusing on the pre-GenAI era with primary technical focus~\citep{adad18,zini22,dwiv23,schw23,rauk23,sae23a,spei22,minh22,bodria23,theissler22,guidotti19,guidotti22} and an interdisciplinary or social science focus~\citep{mil19,mesk22,longo24}. In particular, by building upon them, we perform a meta-survey to structure our methods leveraging also knowledge from pre-GenAI. However, we also uncover novel aspects that have not yet been covered related to GenAI. Many works surveyed various aspects of GenAI (not including XAI) ~\citep{xu23,lin22,xing23,yang23,zhan23b,zhan23,pan23}. We leverage such surveys for our technical background. Some sub-areas of GenAI, e.g., on knowledge identification and editing~\citep{zhan24}, use isolated XAI techniques as a tool, but do not aim at elaborating on it generally. While we could not identify any review discussing XAI for GenAI, some research manuscripts take more of a holistic, partially opinionated view on XAI for large language models (LLMs)~\citep{sing24,liao23} or explicitly survey XAI for LLMs~\citep{zhao23e,luo24}. None of the prior work has provided a comprehensive list of desiderata, motivation, and challenges for XAI for GenAI, and taxonomy. In particular, many of our novel aspects cannot be found in prior works. Aside from that, even when focusing on LLMs only, we differ considerably from prior works.\\

We begin by providing a technical background. To derive the contributions, we proceed as outlined in Figure \ref{fig:out}: Then, we provide motivation and challenges for XAI for GenAI, especially pointing towards novel aspects that emerge with GenAI, such as the increased reach of GenAI throughout society and the need for interactive adjustment of often complex, difficult to evaluate outputs by users. Based on this, we derive desiderata, i.e., requirements that explanations should ideally fulfill, such as the support of interactivity and verification of outputs. Then, we derive a taxonomy for existing and future XAI techniques for GenAI. To categorize XAI we use dimensions related to the in-, the outputs, and internal properties of GenXAI techniques that distinguish them from pre-GenAI such as self-explanation as well as different sources and drivers for XAI such as prompts and training data. 


Using the identified challenges and desiderata, the remainder of this manuscript focuses on discussing novel dimensions for GenXAI and the resulting taxonomy, discussing XAI methods in conjunction with GenAI. Finally, we provide future directions
Our key contributions include describing the need for XAI for GenAI, desiderata for explanations, and a taxonomy for mechanisms and algorithms including novel dimensions for categorization. 

\begin{table}
    \centering
    \caption{Examples of In-/Outputs for GenAI. For more examples, see \citep{goza23}.}
    \label{tab:aisamp}
    \begin{tabular}{p{0.27\linewidth}p{0.41\linewidth}p{0.22\linewidth}}
    \toprule
    Input/Output & Description & Example \\
    \midrule
    \emph{Text to Text} & \begin{tabular}[l]{@{}l@{}}Input: Raw text.\\  Output: Processed or generated text.\end{tabular} & ChatGPT 3.5~\citep{achi23gpt} \\ \hline
    \emph{Text to Image/video} & \begin{tabular}[l]{@{}l@{}}Input: Descriptive text or prompt.\\  Output: Generated image/text.\end{tabular} & DALL-E\citep{bet23}, Sora\citep{vide24} \\ \hline
    \emph{Image/video to Text} & \begin{tabular}[l]{@{}l@{}}Input: Image/video and text.\\  Output: Textual interpretation and answer.\end{tabular} & GPT 4 with Dall-E~\citep{achi23gpt}, Gemini~\citep{reid24}\\ \hline
    \emph{Images, Actions to Actions} & \begin{tabular}[l]{@{}l@{}}Input: Images depicting actions.\\  Output: Generated action sequences.\end{tabular} & Gato~\citep{reed22g} \\ \hline
    \emph{Text to 3D} & \begin{tabular}[l]{@{}l@{}}Input: Text describing object\\  Output: 3D representation of object.\end{tabular} & Magic3d~\citep{lin23mag} \\
    \bottomrule
    \end{tabular}
\end{table}

\section{Technical Background} \label{sec:background}
Here, we provide a short technical introduction to generative AI, covering key ideas on system and model architectures and training procedures. We restrict ourselves to text and image data to illustrate multi-modality. For video and audio, please refer to other surveys (for example \citep{selv23,zha23aud}). 

\subsection{System Architectures}
\begin{figure}[h]
\centering
\includegraphics[width=0.98\textwidth]{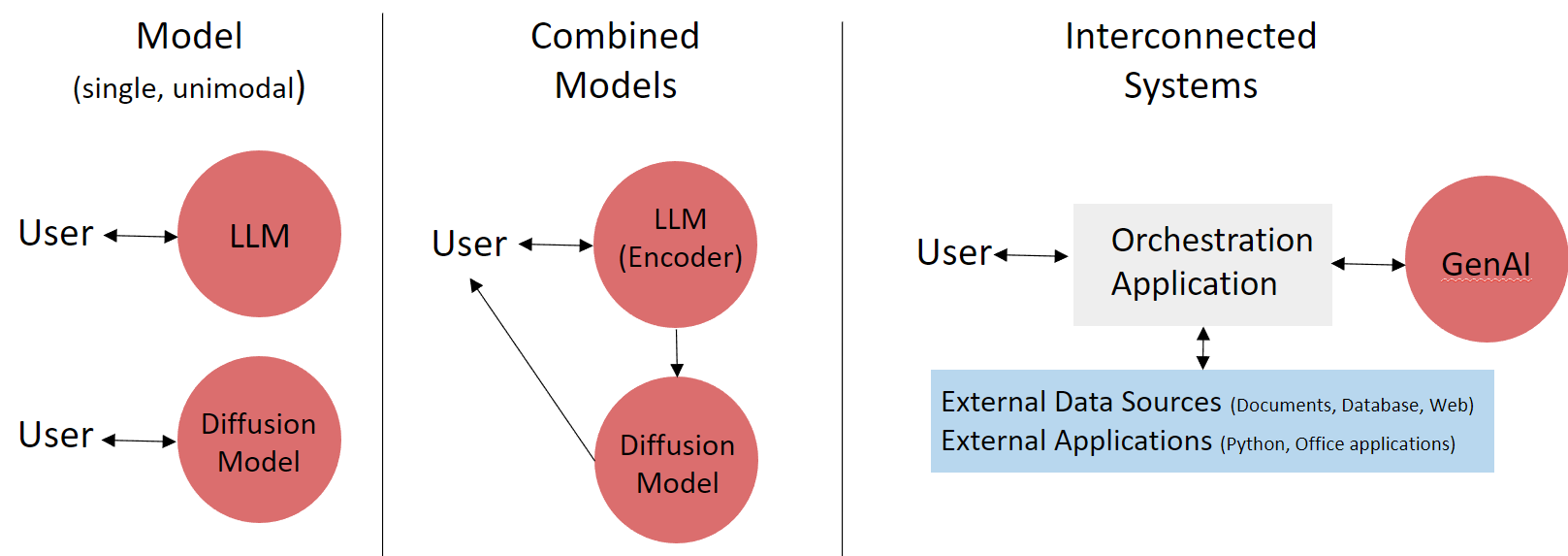}
\caption{Overview of GenAI system architectures comprising a single or combined model (with a GUI) as well as GenAI systems interacting with other applications}\label{lab:mod}
\end{figure}

GenAI models can be used as stand-alone applications accessible through a simple user interface, essentially allowing textual inputs or uploads as for OpenAI's ChatGPT\citep{achi23gpt} and displaying responses. Thus, a system might be essentially one large model, where a model is almost exclusively based on deep learning taking an input processed by a neural network yielding an output. For multi-modal applications, systems that consist of an LLM and other generative models, such as diffusion models, are typically employed. However, GenAI-powered systems might involve external data sources and external applications that interact in complex patterns, as illustrated in Figure \ref{lab:mod}. An orchestration application might decide what actions to take based on GenAI outputs or user inputs. For example, in ChatGPT-4 a user can include a term like ``search the internet'' in the prompt, which implies that first an Internet search is conducted, and retrieved content from the web is then fed into the GenAI model. The orchestration application is responsible for actually performing the web search and modifying the prompt to the GenAI model, e.g., enhancing it with an instruction like ``Answer based on the following content:'' followed by the retrieved information from the web.

\subsection{GenAI Model Architectures}
We discuss key aspects of the transformer architecture and diffusion models and briefly elaborate on other generative models.

\begin{figure}[h]
\centering
\includegraphics[width=0.98\textwidth]{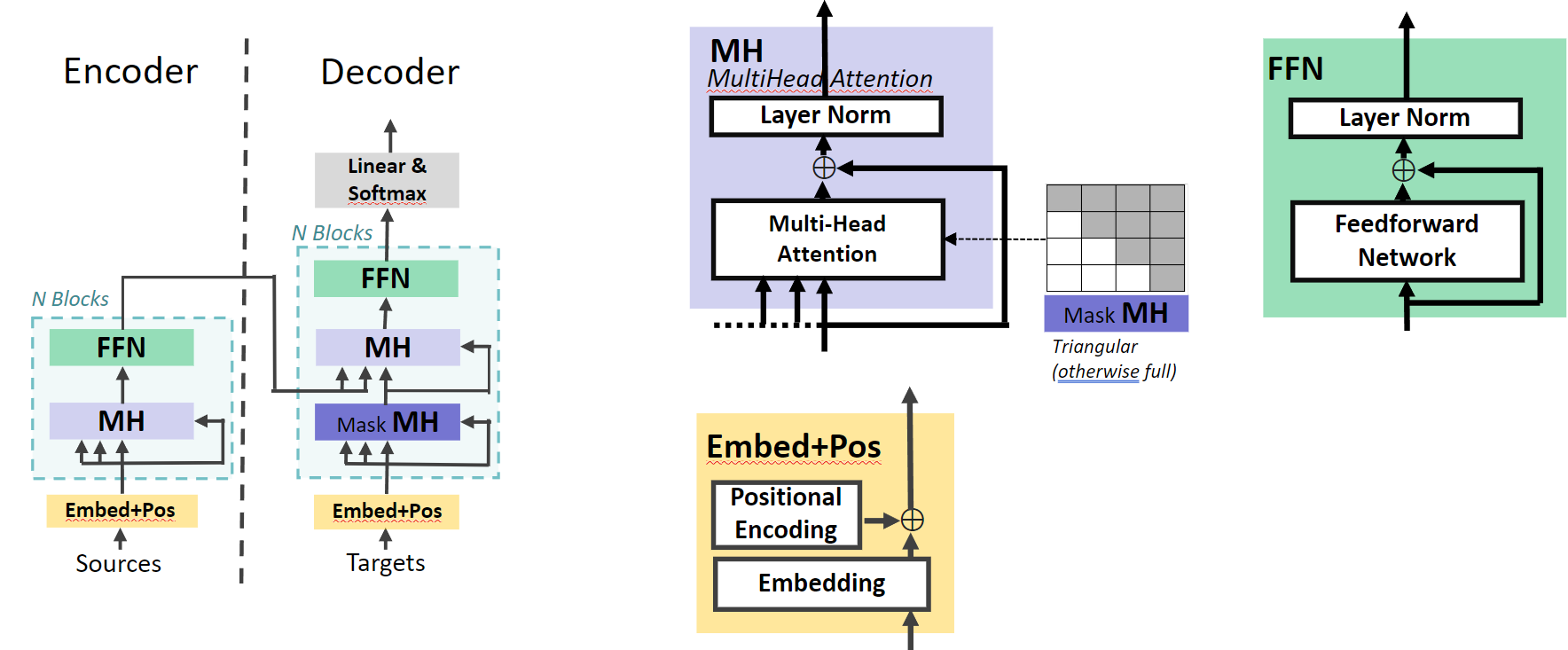}
\caption{Transformer architecture~\citep{vasw17}}\label{fig:tra}
\end{figure}

\subsubsection{Transformers}
Transformers are the de-facto standard for LLMs, while GenAI models involving images might also encompass other models such as diffusion models, variational autoencoders (VAEs), and generative adversarial networks (GANs). The transformer model makes little assumption upon the input data, which makes it a very flexible model. Assumptions on the data (priors) also help to reduce the amount of data needed to train a model. Thus, transformers often require more data than other models to reach the same performance though simpler models might never reach the same top-level performance. The transformer architecture~\citep{vasw17} (Figure~\ref{fig:tra}) comes with many variations~\citep{lin22} mostly by using different implementations of individual elements, for example, using different types of positional embeddings~\citep{duf22} or different types of attention~\citep{de2022a} or even replacing some components, for instance, Hyena~\citep{poli23} provides a drop-in replacement for attention based on convolutions. Commonly, the objectives of these adjustments are better performance and faster computation; for example, the original transformer requires quadratic run-time in the number of inputs, which makes it prohibitive for very long inputs. 
The vanilla transformer architecture (Figure~\ref{fig:tra}) consists of an encoder and a decoder, where the decoder processes the outputs of the encoder and (shifted) targets. One might think of a translation scenario, where the encoder takes a sentence expressed in the source language and the decoder generates one output word after the other in the desired language. Each generated word also becomes an input to the decoder when the next word is generated. Decoder-only architectures (such as the GPT-series~\citep{rad19c,achi23gpt}) do not have an encoder. In contrast, encoder-only architectures do not have a decoder and typically produce contextualized embeddings of single words or text fragments (for example, BERT~\citep{devl18}).

Both the encoder and the decoder take tokens as input. The embeddings are vectors in a latent space. While raw tokens can only be compared for equality, vectors in a latent space allow a more nuanced similarity computation and, potentially, the extraction of specific token attributes such as sentiment. Thus, many current GenAI systems leverage encoders to obtain vector embeddings and use them to retrieve relevant information to enhance prompts by searching in a vector database~\citep{lewis2020retrieval}. A positional encoding is added to the (text) embedding, so the network has information about where a word occurred. After that, inputs are processed through multi-head attention. A single-head attention provides a mechanism to focus (and select) a specific aspect of the input (for instance, syntax or sentiment), which is then further processed through a Feed Forward Network (FFN). Attention can be masked for the decoder so that it cannot access the actual or future targets it wants to predict; for example, for the common task of next-word prediction, the triangular matrix constituting the mask prevents the network has access to the next word to predict as well as words following the next word. 

\subsubsection{Diffusion Models}
Diffusion models learn to reconstruct noisy data. They first distort inputs by repeatedly adding small amounts of noise until the image appears to be noise following, e.g., a Gaussian distribution one can sample from to generate images. They reverse the process for sample generation by taking ``noise'' as input and reconstructing samples. There are multiple rather mathematical intricate methods for diffusion models~\citep{yang23}. Here, we discuss key steps of a prominent technique highly relevant for text-to-image generation, i.e. Denoising Diffusion Probabilistic Model (DDPM)~\citep{ho20}. In the forward pass (input distortion) DDPM acts as a Markov chain, meaning that only the current state (or input) is relevant for the next output, i.e., for a given data distribution $x_0 \sim q(x_0)$, DDPM yields output $x_T$ in a sequence of $T$ sequential steps by computing in step $t$: $q(x_t|x_{t-1})=N(x_t;\sqrt{1-\beta_t}x_{t-1},\beta_t I)$.
Thus, the overall output becomes $q(x_T|x_0)=\prod_{t=1}^T q(x_t|x_{t-1})$. Here $N$ is the Gaussian distribution and $\beta_t$ is a hyperparameter. The reverse pass for generation starts from $p_{\theta}(x_T)$ and yields $p_{\theta}(x_0)$, which is supposed to follow the true data distribution $q_{x_0}$.
Thus, outputs of diffusion models are not easily controllable, i.e., additional input must be provided to the reconstruction process to guide the generation process towards user-desired images as discussed in Section~\ref{sec:con}.

\subsection{Other Generative Models: VAEs, GANs}
We also discuss generative models aside from diffusion models but refer to~\citep{yang23} that performs a detailed comparison for in-depth elaboration. Generative Adversarial Networks (GAN)~\citep{goo24} are trained using a generator that constructs an output from a random vector and a discriminator that aims to distinguish generated outputs $\hat{x}$ from actual samples of the true data distribution $x \sim q_{x}$. An autoencoder~\citep{li2023ae} comprises an encoder and a decoder that learns to output a given input that is compressed by the encoder into a latent space before being attempted to be reconstructed by the decoder. A Variational Autoencoder (VAE) constrains the latent space towards a given prior distribution through a regularization term as part of the optimization objective. If the latent space follows a known distribution, sampling, i.e., sample generation is facilitated.

\subsection{Controlling Outputs} \label{sec:con}
Diffusion and generative models like VAEs and GANs yield high-quality outputs by providing random inputs. Controlling outputs is not easily possible and requires additional effort. Early techniques on controllable image synthesis generated outputs conditioned on class labels or by disentangling dimensions of the latent space so that altering a dimension corresponds to a human interpretable operation~\citep{chen16}. However, these approaches provide only very limited ways to customize outputs. Text-to-image models are more versatile. They commonly encode text and use it as input for generation. Text encoding might be done using a frozen encoder~\citep{saha22} or by altering the text encoder as part of the training process (GLIDE~\citep{nich21}). Text-to-image models require (image, text) pairs for training, in turn, the textual encoder can stem from the text of the (text-image) pairs only or from a broader corpus and larger models. Text-to-image models can either start generation from a low-dimensional space such as Dall-E~\citep{ram22h,bet23} or aim to construct directly from pixel space such as GLIDE \citep{nich21}. For illustration, we briefly discuss Dall-E~\citep{ram22h,bet23}. It uses a multimodal contrastive model where image and text embeddings are matched. It also contains a text-to-image generator before closing the gap between CLIP text and image latent space, which can be learned using a diffusion prior. 

\subsection{LLM Training} 
LLM training consists of at least one and up to three phases. 
The phases differ in the training methodology, goal, and amount of data.

\noindent\textbf{Self-supervised pre-training:} The raw model is trained on very large amounts of data using self-supervised pre-training, for instance, next word prediction (GPT-2) or tasks such as predicting masked words and sentence order (BERT). The goal is to learn a flexible and broad representation of all forms of text that can set the foundation for many different tasks. The data can be a composition of many different kinds of text data as described for LLama-2~\citep{touv23}, e.g., covering code, Wikipedia articles, or a dump of parts of the internet~\citep{Comm}.

\noindent\textbf{Instruction tuning} adapts a pre-trained LLM through supervised fine-tuning~\citep{lou23,zhan23}. The term ``instruction'' is often used interchangeably with prompt. Instructions tend to be more explicit and directive with more precise guidance. The goal of instruction tuning is to improve performance on the most common use cases for LLMs: to follow (short) instructions. Though multiple variations exist, the training data can consist of task instructions, input (task instance), and desired outputs. Instruction tuning helps performance on both seen and unseen tasks and across architectures of different scales~\citep{Longpre}. It can go beyond mere adaption of an LLM and also provide a means to encode additional knowledge for domain specialization, such as medical knowledge~\citep{sing23l}.

\noindent\textbf{Alignment Tuning}: LLMs might not incorporate human values and preferences, i.e., they can produce harmful, misleading, and biased outputs~\citep{shen23l,weid22}. Alignment criteria can be diverse, covering helpfulness, honesty, and harmlessness~\citep{ouya22}. Training data typically stems from humans, for instance, by ranking LLM-generated answers or producing their own answers. Occasionally, also (more powerful) and already aligned LLMs produce training data\citep{alpaca}.  Training can take place by fine-tuning the LLM using supervised learning and the alignment dataset. Training can also be indirect in the sense that human feedback is used to learn a reward model serving as a human critic, i.e., predicting the human scoring of an output. The resulting reward model can be used to adjust the LLM using reinforcement learning. The LLM generates outputs and receives feedback from the reward model. 

\section{The importance, challenges and desiderata of GenXAI}
This section motivates the need to use eXplainable Artificial Intelligence for GenAI and its challenges, followed by describing important desiderata that explanations should fulfill.

\subsection{Importance of XAI for GenAI} \label{sec:und}

\noindent\textbf{Need to adjust outputs - } A larger need for explainability for GenAI emerges as humans effectively control and tailor the generated outputs. Generative AI blurs the boundary between users and developers. It can be instructed to solve tasks based on auxiliary knowledge~\citep{lewis2020retrieval} as witnessed by OpenAI's GPT-4 store~\citep{gptstore23}, which allows ordinary users without programming skills to build and offer applications leveraging uploaded knowledge and GPT-4. 
A new skill of prompt engineering has emerged that users need to master~\citep{zamf23}. XAI has also been identified as a key requirement to support prompt engineering by experts~\citep{mish23}. 
Users need to understand better how to control outputs, handle limitations, and mitigate risks~\citep{weid22}. 
Thus, the stakeholders must understand GenAI to create solutions aligned with their preferences.

\smallskip
\noindent\textbf{Need of output verification - } GenAI, in particular LLMs, are known for many shortcomings, ranging from generating harmful, toxic responses to misleading, incorrect content~\citep{weid22}. 
LLMs can be said to be ``fluent but non-factual,'' which makes their verification even harder. 
Their outputs are generally non-trustworthy and need some form of verification or validation (as also acknowledged by regulators~\citep{EU2023AIACT}). An explanation provides a mechanism to identify errors such as hallucinations in outputs.

\smallskip
\noindent\textbf{Increased reach - } GenAI is easy to access and use through an ordinary browser with a web interface facilitating widespread adoption. ChatGPT was also the fastest product to reach 100 million users and still rapidly grows~\citep{port23g}. It is used throughout society, including vulnerable groups such as schoolchildren, the elderly with limited IT knowledge, and corporate employees.

\smallskip
\noindent\textbf{High impact applications - } GenAI is a general-purpose technology with applications that might have severe immediate and long-term impacts. 
Users might seek advice for pressing personal problems~\citep{sha23} or turn to GenAI for educational purposes. 
In an educational context, using ChatGPT once and receiving a slightly biased response (towards gender, race, or minority) might have limited impact, but receiving such responses over a prolonged time might profoundly impact future generations. 
Aside from such a long-term view, users might turn to GenAI like ChatGPT while being in psychological distress and seeking immediate advice. ChatGPT is even known to outperform humans on emotional awareness~\citep{elyo23}. Given such high-stake applications, understanding GenAI becomes crucial. 

\smallskip
\noindent\textbf{Unknown Applications - } As generative AI can process any text, image, and other medium as input and output, it is hard to anticipate all possible applications. 
As such, the need to understand the models more holistically to ensure they align with higher-level principles becomes relevant.

\smallskip
\noindent\textbf{Difficult to evaluate automatically - } Simple strategies, which means counting the number of correct and incorrect answers (accuracy), provide only a very limited picture of GenAI's behavior, as many tasks yield responses that are difficult to score. 
For example, for summarization, human ratings of gold standards are worse than those of benchmarks, hinting at the challenge in designing benchmarks~\citep{sott23e}. 
Thus, a thorough systematic quantitative evaluation using classical input and output test datasets is difficult, resulting in an increased demand for better model understanding to anticipate potential shortcomings, as automated tests are difficult and possibly insufficient.

\smallskip
\noindent\textbf{Security and safety concerns - } Various problems exist related to the safety and security of GenAI-based technologies, such as witnessed by adversarial examples~\citep{goo24}. 
However, GenAI allows for novel forms of attacks and abuse~\citep{gupt23} targeting a vast number of people through social engineering at a large scale to manipulate elections. 
Also, individuals might leverage GenAI for malicious activities, such as receiving detailed instructions on performing terrorist attacks.

\smallskip
\noindent\textbf{Accountability and legal concerns - } The need for accountability arises due to multiple questions, often driven by legal concerns. 
GenAI exhibits a complex supply chain of data providers and developers, which means data might stem from many sources, and multiple companies might be involved in building a GenAI system. These include those that make a foundation model~\citep{schn24f}, those that fine-tune it to a task, and ultimately, a user who adjusts it through prompting.
This makes accountability a challenging task that is needed as GenAI systems have the potential to cause harm. 
Thus, the question of who is responsible and what causes harm becomes more relevant. 
This can lead to forensic questions like ``Why did an AI trigger a certain action~\citep{sch23t}? 
Was it the training, data, or model?'' 
And even if it can be tied to one aspect, for example, data, further questions come up such as  ``Was it the data from a third-party provider, public data, or data from users?''. The need for accountability arises due to the use of copyrighted material or, potentially, patented material. 
In lawsuits, a key question is if a patent is valid because it is highly original. 
If GenAI can explain that the solution is not ``a copy from an existing patent'' but emerges through basic reasoning given ``prior art,'' the judge might be more inclined to rule the patent invalid.

\begin{table}[]
\centering
\caption{Why is explainability important for GenAI?}
\label{tab:importance_xai}
\begin{tabular}{p{0.2\linewidth}p{0.7\linewidth}}
\toprule
Reason & Description \\
\midrule
\emph{Need to adjust} & Users want to create artifacts aligned with their preferences \\
\emph{Need to verify} & Ensure correctness or identify errors in outputs, for example, due to hallucinations \\
\emph{Increased reach} & Growth in the number and diversity of people, including vulnerable groups exposed to GenAI \\
\emph{High impact applications} & GenAI is used for applications with profound impact on individuals and society, for instance, in the medical domain and education\\
\emph{Unknown Applications} & GenAI can be used for many tasks in ways that are hard to anticipate\\
\emph{Difficult to evaluate} & Many applications of GenAI are hard to evaluate using conventional train/test set approaches, as the quality of outputs is hard to quantify\\
\emph{Security and safety} & GenAI suffers from novel vulnerabilities (such as prompt hacks), and it has the potential to cause harm at a large scale\\
\emph{Accountability and legal concerns} & GenAI involve more actors (data provider, foundation model creator, fine-tuner, end-user), making accountability more challenging while at the same time being subject to legal risks \\
\bottomrule
\end{tabular}
\end{table}

 \begin{table}
\centering
\caption{Why is XAI more challenging for GenAI?}
\label{tab:challenges_xai}
\begin{tabular}{p{0.22\linewidth}p{0.72\linewidth}}
\toprule
Reason & Description \\
\midrule
\emph{Lack of access} & \emph{No access to inner workings} \\
\emph{Interactivity} & Outputs are due to multiple rounds of interaction, necessitating understanding humans and AI and their mutual understanding.\\ 
\emph{Complex systems, models, data, and training} & Understanding gets more difficult as complexity increases, e.g., due to network and training data size, the diversity of data, and the combination of multiple training procedures. GenAI models might also be part of a complex system with multiple closely interacting non-AI components such as code interpreters and search engines.\\
\emph{Complex outputs} & Generated artifacts can comprise millions of bits.\\
\emph{Hard to evaluate explanations} & As automatic evaluation of outputs is hard, so is (function-grounded) evaluation of XAI\\
\emph{Diverse users} & More varied user base \\
\emph{Risk of ethical violations} & As GenAI suffers from biases and offensiveness, so might its explanations\\
\emph{Technical shortcomings} & GenAI models hallucinate and exhibit reasoning errors. In particular, also self-explanations by LLMs are subject to known shortcomings such as hallucinations and limited reasoning \\

\bottomrule
\end{tabular}
\end{table}

\subsection{Why is XAI more challenging for GenAI?} \label{sec:cha}
\noindent\textbf{Lack of access}: Commercial GenAI models from large corporations such as OpenAI and Google are among the most widely used systems. However, users, researchers, and other organizations interested in understanding the generative process of specific artifacts or models cannot access model internals and training data of these models. This rules out many XAI approaches.

\smallskip
\noindent\textbf{Interactivity}: Some tasks involving humans and GenAI are inherently interactive, including negotiations that can be conducted between GenAI and humans~\citep{sch23neg}. Thus, explanations need to focus not just on the model, but also on how the model impacts humans and vice versa over the course of an interaction.

\smallskip
\noindent\textbf{Complex systems, models, data, and training}:  Understanding AI gets increasingly difficult as models grow and digest more training data. GenAI based on very large foundation models constitutes the largest AI models today with hundreds of billions of parameters~\citep{schn24f}. They also lead to novel, more complex supply chains of AI. Pre-GenAI models were often built based on company-internal data, possibly by fine-tuning a model trained on a moderate-sized public dataset such as ImageNet. GenAI systems are built using much larger data from many sources, including public and third-party providers. Often, a foundation model is further adjusted through fine-tuning or retrieving information from external data sources~\citep{lewis2020retrieval} to yield a GenAI system. Thus, the final output of a GenAI system might be generated not only through a deep learning model but also by engaging with other tools such as code interpreters~\citep{schi24t}.

\smallskip
\noindent\textbf{Complex outputs}: Generated artifacts are complex, i.e. typically consisting of thousands up to millions of bits in an information-theoretic sense. Classical supervised learning yielding classes often produce just a single bit in case of binary decisions and at most a few dozen bits, i.e., mostly less than a few million classes. In some sense, before GenAI a label of a classifier constitutes a single decision, while GenAI makes a multitude of decisions -- one for each aspect of the artifact. Thus, textual outputs and images naturally lead to investigations of several aspects of the output, such as tone, style, or semantics of texts~\citep{yin22}. There are many possible questions about why an artifact exhibits a certain property.

\smallskip
\noindent It is \textbf{hard to evaluate explanations}, especially for function-grounded approaches, where GenAI itself is hard to evaluate. Function-grounded evaluation focuses on assessing an XAI method based on predefined benchmarks, for example, to use an explanation to classify an object and see whether the explanation is equal to the output of the model used to create the explanation. As such benchmarks might not exist and are difficult to create for certain GenAI tasks, evaluation of explanations also becomes more challenging.\\

\smallskip
\noindent\textbf{Diverse users}: A model might be utilized by a diverse set of users across all age and knowledge groups covering many needs -- similar to a search engine. In contrast, pre-GenAI systems are more often tailored to a specific user group and task.

\smallskip
\noindent\textbf{Risk of ethical violations}: Even commercial GenAI models are known for producing potentially offensive, harmful, and biased content~\citep{achi23gpt}. GenAI models also commonly self-explain, and, in turn, the possibility arises that while the outputs might be ethical, the explanations are offensive, for example, due to inadequate phrasing or visual depictions. 

\smallskip
\noindent\textbf{Technical shortcomings}: GenAI suffers from hallucinations and limited reasoning capability. In turn, if GenAI models self-explain, such explanations are subject to the same shortcomings.

\subsection{Desiderata of Explanations for GenAI} \label{sec:desi}
\begin{table}[htbp]
    \centering
    \begin{tabular}{p{2.5cm} p{9cm}}
        \toprule
        \textbf{Novel Desiderata} & \textbf{Description} \\
        \midrule
        \emph{Verifiability} & Explanation should support verification of AI outputs.\\
        \emph{Lineage} & Explanations might include information on tracking and documenting data, algorithms, and processes in AI models to aid accountability, transparency, and reproducibility. \\
        \emph{Interactivity and personalization} & With complex AI systems, explanations must be customizable by users and their preferences. \\
        \emph{Dynamic explanations} & Explanations might differ in content and structure depending on the sample and specified objectives (e.g., max. plausibility, explain surprising aspects) as it is impossible to explain all details of complex artifacts.\\
        \emph{Costs} & Economics of XAI include costs associated with risks of value leakage and the costs for implementing XAI. \\
        \emph{Alignment criteria} & Explanations should align with criteria like helpfulness, honesty, and harmlessness, crucial for trust and ethical AI. \\
        \emph{Security} & Providing explanations should not compromise security, as insights on GenAI models might facilitate attacks or exploit vulnerabilities.\\
        \emph{Uncertainty} & Understanding and conveying the confidence and uncertainty of AI outputs is vital for decision-making and trust.\\
        \bottomrule
    \end{tabular}
    \caption{Overview of novel and emerging desiderata for GenXAI}
    \label{tab:summary}
\end{table}

\noindent There are also several \textbf{novel and emerging aspects} as well as a big change in the relevance of desiderata:
\begin{itemize}
    \item \textbf{Verifiability} has been recently discussed as an important aspect of explanations~\citep{fok23}. But the problem that outputs of LLMs cannot be verified (due to hallucinations) has been discussed years earlier~\citep{may20}. Suppose explanations cannot be verified, and an explainee must trust the explanation. In that case, the consequence can be the rejection of a correct answer due to an incorrect explanation or failure to detect incorrect outputs. A key concern concerning verification is the effort to verify. Efforts to understand and tailor explanations have been discussed for XAI in general, for example, for time efficiency see~\citep{schw23}.
    
    \item \textbf{Lineage} ensures that model decisions can be tracked to their origins. It refers to tracking and documenting data, algorithms, and processes throughout the lifecycle of an AI model. It is highly relevant for accountability, transparency, reproducibility, and, in turn, governance of artificial intelligence~\citep{sch22gov}. It concerns the ``who'' and the ``what'', for example, ``Who provided the data or made the model?'' or ``What data or aspects thereof caused a decision?''. While the latter is a well-known aspect of XAI, as witnessed by sample-based XAI techniques, the former has not been emphasized significantly in the context of XAI. ~\citep{fau23} set forth data traceability as a requirement in the context of Machine Learning Operations (MLOPs) for XAI in industrial applications. The need for lineage emerges as GenAI supply chains get more complex often involving multiple companies~\citep{schn24f} rather than just a single one. Furthermore, multiple lawsuits have been undertaken in the context of generative AI, for example, related to copyright issues~\citep{Gry23}. Regulators have also set stringent demands on AI providers~\citep{EU2023AIACT}. Thus, employing GenAI poses legal risks to organizations. In turn, ensuring lineage-supported accountability can serve as a risk mitigation.
    
    \item \textbf{Interactivity and personalization} have been discussed previously in XAI~\citep{schw23,sch19p}. However, as GenAI outputs and systems are more complex, the number of options for how and what to explain has increased drastically. While there is only one bit to explain in a binary classification system, it amounts to millions for a generative AI system generating images. That is, it is nearly impossible for a user to understand the reasons behind all possible details of an output. Thus, depending on user preferences and the purpose of the explanation, some aspects might be more relevant to understand than others, necessitating the need for a user to engage with the system to obtain explanations and for the system to provide adequate explanations tailored to the user's demand. As discussed in Section~\ref{sec:inter}, systems supporting interactive XAI are increasingly emerging.

    \item{Dynamic explanations} aim at choosing the explanation qualities and content (e.g., What output properties are explained?) automatically depending on the sample, explanation objective and possibly other information. For example, an explanation might elaborate on why a positive or negative tone for a generated text is chosen. Tonality might only be discussed for some samples and not for others. When to include it, should be aligned with a specified objective, e.g., the explanation should satisfy the explainee (plausibility) or to explain the most surprising aspects of the generated artifacts or to explain the most impactful properties of the artifact on its consumers. The goal of dynamic explanations is to maximize the specified objectives while being free to choose the explanation content, including meta-characteristics such as what aspects are included in the explanation, its structure, etc.    
    
    \item \textbf{Costs} related to XAI might also become an emerging area. Economics of XAI has been elaborated in~\citep{beau20}, where concerns related to the costs of implementing XAI (and transparency) are mentioned. For cooperations, GenXAI adds the risk of leaking value. That is, competitors (or academics) might prompt a model to save on the costs of training data. For example, the Alpaca model~\citep{alpaca} was trained on data extracted from one of OpenAI's models, allowing them to forego the costly and time-consuming task of collecting data from humans. Using explanations, for instance, as part of chain-of-thought (CoT)~\citep{wei2022chain}, can further enhance performance and is thus of value.
    
    \item \textbf{Alignment criteria} such as helpfulness, honesty, and harmlessness~\citep{ask21} that are relevant for GenAI in general also play a role for XAI. Some of these criteria partially overlap with existing ones, such as plausibility (with helpfulness) and faithfulness (with honesty, for instance, explanations should not be deceptive~\citep{schn23dec}). Aspects such as harmlessness have received less attention, as explanations were commonly simpler (for example, attribution-based). Harmlessness implies that explanations do not contain offensive information or information concerning potentially dangerous activities~\citep{ask21}. It also relates to security, discussed next: 
    
    \item \textbf{Security} increasingly evolves as a desideratum. Explanations should not jeopardize the security of the user and the organizations operating the GenAI model. Providing insights into the reasoning process might facilitate attacks or might be leveraged in competitive situations, leading to poorer outcomes of GenAI. For example, in a recent study on price negotiations of humans against LLMs~\citep{sch23neg}, humans asked an LLM what decision criteria it used for its decisions and, in turn, systematically exploited this knowledge to obtain better outcomes against the LLM. A human negotiator might not disclose such information. Thus, openness can also be abused. ``Security through obscurity'' is one protection mechanism against attacks. Cooperation might also aim to protect their intellectual property. For example, customer support employees (and GenAI models) might have access to some relevant information on a product to help customers. Still, they might not be allowed to obtain explanations on how exactly the product is manufactured as this might constitute a valuable secret for the company. 
    
    \item \textbf{Uncertainty}: Understanding the confidence of outputs is an important desideratum of XAI. While most works aim to explain a decision, explaining the uncertainty of a prediction has also attracted some attention~\citep{mol20,gawl23}. Deep learning models such as image classifiers are known to be overconfident, with multiple attempts to address the issue~\citep{mero24,gawl23}. LLMs arguably take this a step further, as they commonly generate answers eloquently even if they are wrong, i.e., they are ``fluent but non-factual''. Still, LLMs have some (though not perfect) understanding of whether they can answer a question~\citep{kada22} or not. They can also be enhanced with uncertainty estimation techniques~\citep{huan23}.
\end{itemize}


\noindent Prior research has extensively discussed principles and desiderata of explanations. Here, we briefly summarize key characteristics explanations should exhibit drawing on prior surveys~\citep{lyu24,sch19p,schw23,bodria23,guidotti22} and briefly discuss based on novel characteristics of GenAI. Among the most important, well-known desiderata are:

\begin{itemize}
    \item \textbf{Faithfulness} (= fidelity = reliability): An explanation should precisely reflect the reasoning process of a model. Due to the complexity of GenAI (models), a higher level of abstraction seems necessary to keep explanations comprehensible in a limited amount of time.
    \item \textbf{Plausibility} (= persuasiveness = understandability): An explanation should be understandable and compelling to the target audience. Textual explanations (especially, self-explanations) are often easy to understand (e.g., compared to classical explanations such as SHAP values).
    \item \textbf{Completeness} (= coverage) and minimality: An explanation should contain all relevant factors for a prediction (completeness) but no more (minimality). Asking for complete coverage becomes less feasible with the growth of output, data, and model sizes. Personalized, interactive explanations allowing a user to control explanations as well as XAI techniques automatically selecting only interesting properties to be explained could be the way forward.    
    \item \textbf{Complexity}: Total amount of conveyed information in an explanation, typically measured relative to an explainer's knowledge, i.e., subjectively. This aspect gains in importance as stakeholders become more diverse.
    \item \textbf{Input and model sensitivity and robustness}: Changes in the input (or model) that impact model outputs should also lead to changes of explanations (sensitivity), but if changes (of inputs) do not alter model behavior or changes in the model do not alter processing and outputs significantly then explanations should not change disproportionately (robustness). This still holds for GenAI.
\end{itemize}

\begin{table}[h]
\centering
\caption{Dimensions of our taxonomy for GenXAI algorithms}
\label{tab:xai_hierarchy}
    \begin{tabular}{@{}p{0.99\linewidth}@{}}
        \hline
        \textbf{Dimension} \\
        \hline
         \emph{\textbf{Explanation (=Output) properties:}} \\
         \textbf{$\triangleright$ Scope} \\
        \begin{tabular}{@{\hspace{0.5cm}}l}
            $\circ$ Explaining single vs. all attributes of the output \\
            $\circ$ Explaining single input-output vs. entire interaction \\
        \end{tabular} \\
        \textbf{$\triangleright$ Modality} \\
        \begin{tabular}{@{\hspace{0.5cm}}l}
            $\circ$ Unimodal vs. multi-modal explanations\\
        \end{tabular} \\         
        \textbf{$\triangleright$ Dynamics} \\        
        \begin{tabular}{@{\hspace{0.5cm}}l}
            $\circ$ Interactive vs. non-interactive explanations\\
            $\circ$ Static (sample-independent) vs. dynamic (sample-dependent) explanation qualities and content\\       
        \end{tabular} \\         
        \smallskip        
        \emph{\textbf{Input and internal properties:}} \\        
        \textbf{$\triangleright$ Foundational source for XAI} \\
        \begin{tabular}{@{\hspace{0.5cm}}l}
            $\circ$ Model\\
            $\circ$ Optimization\\
            $\circ$ Training data\\
            $\circ$ Prompt\\
        \end{tabular} \\
        \textbf{$\triangleright$ Required access by XAI Method} \\
        \begin{tabular}{@{\hspace{0.5cm}}l}
            $\circ$ Black-box vs. white-box\\
        \end{tabular} \\
        \textbf{$\triangleright$ Model (self-)explainers} \\
        \begin{tabular}{@{\hspace{0.5cm}}l}
            $\circ$ Self- vs. explanations by other models/algorithms \\	
        \end{tabular} \\
        \textbf{$\triangleright$ Sample difficulty} \\
        \begin{tabular}{@{\hspace{0.5cm}}l}
            $\circ$ Explaining simple vs. difficult samples \\	
        \end{tabular} \\
        

        \smallskip
        \textbf{$\triangleright$ Dimensions of pre-GenAI} \\
        \hline
    \end{tabular}    
\end{table}

\section{Taxonomy of XAI techniques for GenAI}
Our taxonomy provides a scheme for the classification of XAI mechanisms and algorithms supporting the understanding of GenAI (Section \ref{sec:dim}), which we utilize to classify existing techniques (Section \ref{sec:class}).

\subsection{Dimensions of taxonomy} \label{sec:dim}
The key characteristics of our taxonomy are summarized in Table~\ref{tab:xai_hierarchy}. We distinguish between GenXAI algorithms' output (= explanation) properties, input, and internal properties. Explanation properties characterize the XAI algorithms outputs, i.e., the explanations, in terms of scope, i.e., what (fraction of) samples, attributes and part of the interaction they explain, modality, that means, unimodal or multi-modal, and interactivity, i.e., can user engage in obtaining additional explanations or tailor explanations. Input and internal properties relate to what the XAI algorithms require producing explanations and how they are obtained.
While many ideas on structuring XAI are still valid in the context of GenAI, we focus primarily on novel dimensions such as the foundational source for XAI, which is one of data, model, training, and prompt. The source forms the key mechanism or artifact leveraged by XAI techniques to generate explanations, as elaborated in Section~\ref{sec:fou}. One might classify the first three (data, model, and training) under the category of intrinsic methods as they impact the resulting model. However, data can also be extrinsic, particularly for Retrieval-Augmented Generation (RAG). Similarly, for prompts, Chain-of-Thought (CoT) encourages and guides XAI, but it also relies to some extent on training. That is, if training data lacks any form of explanation, CoT prompting will not work. This can be most easily seen in an extreme case, where words like ``because'' are removed from the training data, implying that they will never be generated. 

\subsubsection{Output, Interaction and Input Scope} \label{sec:scope}
\begin{figure}[h]
\centering
\includegraphics[width=0.9\textwidth]{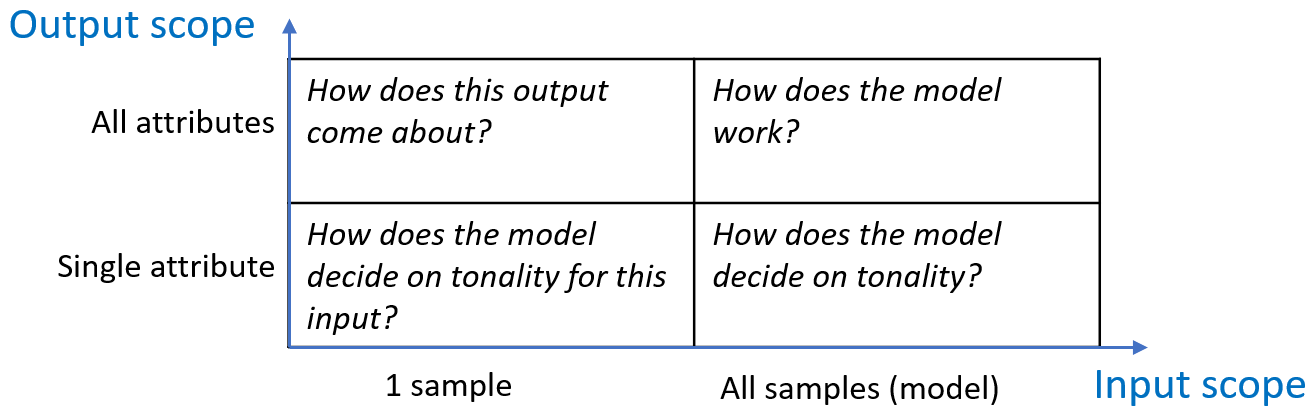}
\caption{Illustration of input and output scope, where we use ``tonality'' as an example of a text-related attribute. Traditionally, scope only referred to output scope.}\label{fig:inout}
\end{figure}

Often, scope~\citep{schw23} only refers to what inputs are explained by a method, i.e., a single sample (local) versus all samples (global), which means the model behavior in general~\citep{guidotti19,bodria23}. Thus, the scope states the quantity of the input samples, which are explained. We call it \emph{input scope}.
For GenAI, we say that scope also refers to the quantity of the output that is explained, i.e., a single attribute of the output (focused) vs. all attributes (holistic). We call this \emph{output scope}. This is illustrated with an example in Figure~\ref{fig:inout}. As outputs are significantly more complex for GenAI, they raise more options for questions. For example, why did the response contain some information and not another? Why was the sentiment of a generated sentence positive, neutral, or negative? While some of our methods touch on these questions, overall, there is limited work in this direction. Similarly, there is limited understanding of how to relate training data to predictions beyond classical approaches such as influence functions, i.e., how did a specific piece of knowledge in the training data impact the generation process?  

\begin{figure}[h]
\centering
\includegraphics[width=0.9\textwidth]{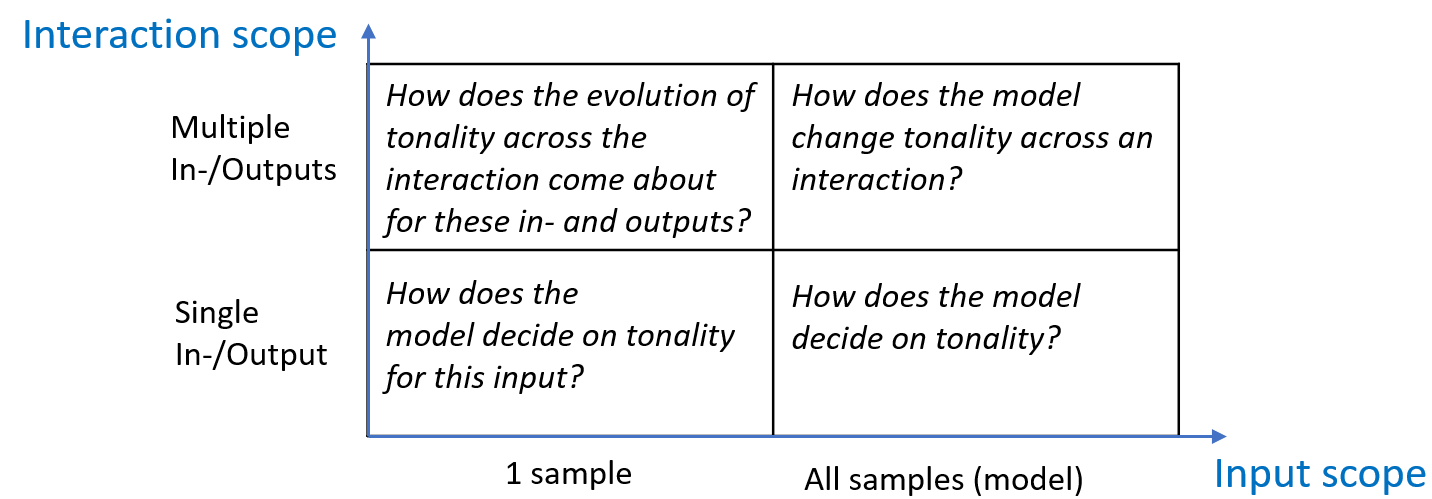}
\caption{Illustration of input and interaction scope, where we use ``tonality'' as an example of a text-related attribute. Traditionally, scope only referred to output scope.}\label{fig:inter}
\end{figure}

\noindent \emph{Interaction scope}: For GenAI, interactions are no longer commonly in the form of dialogues, and understanding both sides of the system becomes much more relevant. As such, we distinguish two goals:
\begin{itemize}
    \item  Explaining (single) input-output relations: This is the classical notion, where an AI system processes one input to produce one output. The goal can be to understand a particular instance (local explanation) or the model as a whole. While explanations might consider user-specific aspects and personalize explanations~\citep{sch19p}, such personalization is typically independent of the interaction. 
    \item  Explaining the entire interaction: It focuses on human-AI interaction and its dynamics, for example, the communication and actions between an AI and a human when a human solves a task using an AI. Interactions are characterized by multiple rounds of outputs generated by both sides conditioned on prior outputs. In this case, the goal is to more holistically explain \textit{(i)} the dynamics of the interaction, i.e., not just one single input-output but the entire sequence of in- and outputs and \textit{(ii)} the outcome of the interaction, which could be why a particular artifact such as an image was generated in a certain way, but also why a user did not complete a task, for example, abandoned it prematurely or achieved an unsatisfactory output.
    Interaction dynamics are influenced by a series of technical (such as model behavior including classical performance measures but also latency, user interface, etc.) and non-technical factors (such as human attitudes and policies). As such, human-AI interaction cannot easily be associated with one scientific field but is inherently interdisciplinary. Explainability, which aims at understanding AI technology, should focus on how technical factors related to model behavior impact the interaction. While many existing works touch on the subject, the change in interactivity brought along by prompting due to GenAI is not well understood. A study that investigated interactivity in negotiations was~\citep{sch23neg}. However, the explanations for outcomes of negotiations and interaction behavior are not through algorithms but rather through a manual, qualitative investigation of interaction -- a common technique in social sciences but less prevalent in computer science.
\end{itemize}
The two goals are illustrated with an example in Figure~\ref{fig:inter}. 

The field of human-computer interaction has aimed at explaining human interactions for a significant amount of time~\citep{mack24,carr88} with some effort also on discussing human-AI interaction. For example, guidelines for human-AI interaction (before GenAI) are well-studied~\citep{ame19}, also the topic of explanation in collaborative human-AI systems~\citep{the23}. Furthermore, explanations in human-AI systems typically encompass objectives that are driven by non-technical concerns~\citep{mil19,mue21} but there is less work on explaining human-AI interactions themselves~\citep{sree22}, in particular, targeted towards GenAI. To provide two examples from the pre-GenAI area spanning from in-depth technical study to broader organizational studies: \citep{sch22opt} discussed how human-AI interaction could be optimized, accounting for long-term goals such as preserving human diversity. The work would explain how the user can improve her interaction to reduce error rates and become more efficient. \cite{gris23} investigated the dynamics of AI adoption within an organization, explaining how error rates and learning behavior of an AI impact the complexity of processes. Explanations in interactive systems in pre-GenAI~\citep{rag21} and GenAI (see Section~\ref{sec:inter}) are typically more concerned about supporting interactive explorations of a decision, for instance, querying a system to better understand it rather than obtaining explanations on how a sequence of inputs and outputs emerged.

\subsubsection{Explanation Modality} \label{sec:multimodal} 
Commonly, explanations are unimodal, for example, textual, visual, or numeric. However, also multi-modal explanations have been investigated~\citep{park18} by collecting a dataset containing textual and visual justifications. They show that multi-modal explanations yield favorable outcomes, for instance, each modality improves due to the presence of the other.

\subsubsection{Dynamics} \label{sec:inter}
\noindent \emph{Interactivity:} In a classical setting, XAI is often non-interactive, meaning an explainee has limited options to control explanations or request additional information. However, the idea of interactivity has existed in XAI for some time, as shown in prior taxonomies~\citep{schw23}. One idea is to rethink XAI as a dialogue~\citep{sing24}. \cite{slack23} supports the explainability of a machine learning model by using an LLM to translate natural language queries into a pre-defined set of operations related to explainability, for example, getting the most important features, changing a feature, etc.. \cite{gao23} uses LLMs to create an interactive and explainable recommender system. In terms of classical metrics such as recall and precision, the system is not outperforming. Improving models due to human explanations in an interactive setting has also been studied for image recognition~\citep{schra20}.
While LLMs are known for their superior performance in many Natural Language Processing (NLP) tasks, they might also be employed without improving classical metrics such as accuracy or precision and recall but rather to yield other benefits such as explainable, interactive systems requiring less (training) data~\citep{gao23,deld23}. \cite{deld23} showed that binary risk classification can be done with 40x fewer data to roughly match the performance of a classical machine learning system through explainable-guided prompts.\\ 
\noindent \emph{Static vs. dynamic explanation qualities and content} relates to how the structure and content of explanation vary based on the sample due to the algorithm. That is, which and how properties of the generated artifact are explained. This choice can be independent of the sample(static), i.e., identical for each sample, or it can be dependent on the sample itself(dynamic). For example, classical methods like attribution maps are static, as they only explain relevance and they always assign a relevance value for each input pixel. In contrast, textual self-explanations can be dynamic. The explained properties can be selective, e.g., for a generated image an explanation like ``A moving, red car was shown to make the image more vivid and engage viewer'' only describes one part of the image and one property of that part. Explanations are dynamic if the structure and content of the explanation varied with the sample. Say the XAI technique yielded a generated image of a mountain landscape ``A snowy landscape was chosen as the prompt contained the word ``bright''. Explanations were dynamic as the two exemplary explanations refer to a different object (car vs landscape), differ in explained properties (explaining emotional intention vs. not discussing it), and differ in causal factors discussed (mentioning what inputs caused output properties, i.e., ``bright'' in prompt vs. not doing so).

\subsubsection{Foundational Source for XAI techniques}\label{sec:fou}
We consider the data, model, and optimization (training) and prompt the foundational sources for XAI methods. That is, each source can be modified or tailored to improve XAI. 

\smallskip
\noindent\textbf{Model} induced XAI refers to intrinsic XAI methods (also named model-specific XAI~\citep{guidotti19}), where the design of the model is altered to foster explainability. With GenAI, novel intrinsic methods have emerged.\\
\emph{Using interpretable components}: Deep learning is a composition of multiple layers and components, such as activation functions and attention mechanisms. These components can be more or less complex, limiting the overall explainability of the model.
For example, interpretable activation functions might substitute conventional less interpretable functions, for instance, SoLU~\citep{Elhage2022SOLU} is said to enhance interpretability. These functions are built on the idea that there are more features than neurons (per layer), and, in turn, superposition yields additional features~\citep{olah20}. The paper also provides evidence for this hypothesis. The interpretability makes some neurons more interpretable but hides others. Thus, while the method overall claims to provide a net benefit, it is not without costs.
Classical attention layers are also commonly used, though their value for XAI is debated (see Sec.~\ref{sec:Attr}).

\smallskip
\noindent\emph{Interpretable GenAI models through additional models}: Model interpretability can be achieved by combining LLMs with other models and often external knowledge. For example, in~\citep{chen23e} an LLM is enhanced with a GNN and external knowledge that generates an explanation and prediction jointly. \cite{cres22} fostered explicit multistep reasoning by chaining responses of two fine-tuned LLMs; one performs selection, the other inference. \cite{wang23e} uses the idea to incorporate external knowledge using classical internet search (as is done in commercial products such as Bing Chat). In addition, it also uses first-order logic; for example, it creates easier-to-verify subclaims (that jointly lead to the overall claim). It also generates explanations simply by querying the LLM. \cite{vedu23} developed and trained a decoder for faithful and explainable online shopping product comparisons.

\smallskip
\noindent\textbf{Optimization}: Adjusting the optimization objective is a common technique to foster explainability. Common strategies involve disentanglement of latent dimensions~\citep{ross2017right} as well as training also for XAI relevant criteria, such as citing evidence~\citep{meni22}. In the context of GANs, \cite{chen16} trained explicitly disentangled representations by maximizing the mutual information between a small subset of the latent variables and the sample. This allowed the so-called InfoGAN to separate writing styles from digit shapes on the MNIST dataset. Furthermore, in~\citep{ross2017right}, explanations have been used to constrain training, such as ``to be right for the right reasons''. Note that while disentanglement is commonly achieved through training objectives and regularization, special network architectures like capsule networks~\citep{sabour2017dynamic} and others (see Section on Disentanglement in~\citep{schw23}) can also support disentanglement. 
Diffusion models already have a semantic latent space~\citep{kwon22}. A special reverse process can leverage it for image editing using CLIP~\citep{ram22h}, which iteratively improves reconstructed images. 
\citep{meni22} trains (more than one) model to cite evidence for claims, facilitating answer understanding and, especially, fact-checking.

\smallskip
\noindent\textbf{Training data} can support XAI in multiple ways. While the exact impact of training data and dynamics is not yet fully explored~\citep{teeh22}, current findings indicate that training data is an important factor to support explainability -- particularly for textual data.\\
\noindent\emph{Training data composition:} GenAI training data is usually complex -- even for a single modality. For example, text data might consist of programming code, books, dialogues, etc. 
The composition of the training data can impact reasoning, i.e., code can improve certain reasoning task~\citep{ma23}. While less explicit statements are known for explanations, the overall data composition is also likely impacting XAI.\\ 
\noindent\emph{Explanation quality and quantity}: Aside from the overall composition of training data, the presence and absence of explanations in the training data also impacts XAI. While it is well-known that the quality of training data has a strong impact on the performance of a model, the impact on XAI has received less attention. However, as GenAI can self-explain as witnessed by CoT~\citep{wei2022chain}, i.e., generate explanations as part of outputs, training data strongly impacts performance. For illustration, assume that a model is trained with erroneous explanations though potentially correct results, it could in principle perform well on tasks but provide poor explanations. Also, if the training data does not contain any data explaining the reasoning, the model might perform worse at explanations. In an extreme case, if the training data does not contain explanations including explanatory words such as ``due to'', ``for that reason'' and ``because'' the model will also never generate such words (and the corresponding explanations).\\ 
\noindent\emph{Domain specialization}~\citep{ling2023domain} can tailor the model more towards a specific domain, potentially at the cost of abilities in other domains~\citep{chen20rec}. It can also contribute towards explainability, as a narrower model focus implies fewer potential options for explanations and, thus, a lower risk of errors and easier verifiability.\\ 

\smallskip

\noindent\textbf{Prompts} can also induce explanations: LLMs can be prompted \textit{(i)} to provide explanations in a preferred manner or \textit{(ii)} be constrained to rely on a limited set of given facts to create responses, which facilitates verification. The idea of ``rationalization'' dates back to the early 2000s~\citep{zaid07}. One type of explainability is ``justifying a model's output by providing a natural language explanation''~\citep{gurr23}.
Commonly, explanations clarifying the reasoning process are elicited through the use of chain-of-thought(CoT) prompting~\citep{wei2022chain}. Such prompting allows the structure of the reasoning process, implicitly shaping explanations and utilizing external knowledge for each reasoning step to yield more faithful explanations~\citep{wei2022chain,he22}. However, explanations can also contain hallucinations, as shown in the context of few-shot prompting~\citep{ye2022un} and sensitivity to inputs for CoT prompting~\citep{turp24}. Despite their unreliability (to explain the true inner workings due to lack of faithfulness), they might still be of value to verify output correctness or, in the training of (smaller) models~\citep{zhou2023flame}.
Additionally, LLMs can be constrained to rely on information, which is part of the user-given prompt or extracted from an external database or the internet, known as Retrieval-Augmented Generation (RAG)~\citep{lewis2020retrieval}. RAG facilitates understanding the response of an LLM and, especially, verifying it, as the source of information for the answer is known and typically very small compared to the entire training data of a GenAI model.
To facilitate explainability for recommender systems, personalized prompt learning, for example, soft-prompt tuning to yield vector IDs, has been conducted~\citep{li23p}.


\subsubsection{Required Model Access by XAI Method}
Depending on what is to be explained and the XAI method, different information is needed to obtain explanations~\citep{schw23}. Black-box access does not provide information on likelihoods other than the predicted output, and potentially even such probabilities might not be given. That is, internal access, e.g., to activations and gradients, is commonly unavailable. White-box access refers to having complete access to the model, its training data, and its training procedure. In between, there is a wide range of grey-box access~\citep{sch23t}.
An important restriction for XAI techniques investigating commercial models is that models are typically only accessible as a black box, such as GPT-4 through an API.  Similarly, commercial vendors do not share their training data and often do not even disclose a coarse summary of the training dataset. Thus, also XAI techniques leveraging training data are not easily employable.

\subsubsection{Model (Self-)Explainers} 
GenAI, in particular, LLMs provide explanations for their own decisions (self-explain) or serve as explainers in general. That is, the model itself provides explanations rather than a dedicated XAI technique. This contrasts the classical notion of intrinsic XAI that often denotes understandable, simple models such as decision trees and linear regression.
While self-explanations are not necessarily faithful~\citep{turp24}, attempts have been made to improve them. For example, \cite{chua24} proposed to use an evaluator to quantify faithfulness and optimizing faithfulness scores iteratively. Though self-explanations are not necessarily accurate or faithful, they can be helpful, as demonstrated in a complex environment, where agents performed multistep planning and improved through self-explanation~\citep{wang23}.

LLMs can serve themselves as explainers by providing explanations for essentially anything, for instance, they can self-explain by generating explanations tailored to their outputs~\citep{wei2022chain,turp24} or support explaining other machine learning models~\citep{openai_2023b,slack23,sing23}, provide explanations by analyzing data, for example, pattern in data through autoprompting~\citep{sing22}, support people in self-diagnosis~\citep{sha23}, or yield interpretable autonomous driving systems~\citep{mao23}. In fact, on free-form coding tasks (generation), LLMs produce explanations that often exceed the quality of crowd workers’ gold references~\citep{ziem23}. For mental health analysis, explanations of LLMs approach those of humans in quality~\citep{yang23t}. 


\subsubsection{Explanation Sample Difficulty} \label{sec:expdi} 
Not all input samples are equivalently challenging to explain~\citep{saha22h}. The idea that some samples and interactions are easier to explain than others gains in relevance, as there is a larger variance in possible inputs and outputs. For example, LLMs allow us to ask the most simple to the most complex questions from a human perspective. LLM-generated texts might be a ``lookup'' of a fact learned from the training data onto long stories or solutions to complex tasks.  Explanations (as judged by humans) might be worse for difficult samples than for simple ones. More precisely, this has been observed in explaining data labels with GPT-3, where GPT-3 explanations degraded much more with example hardness than human explanations~\citep{saha22h}. Generally, the idea of distinguishing XAI tasks based on their difficulty has received little attention so far. From a computational perspective, current explanation algorithms require an identical amount of computation independent of the difficulty. Also, forward computations typically proceed identically independent of sample difficulty, though the notion of reflection (and thinking fast and slow) has been mentioned in the literature~\citep{sch23ref}. LLMs can benefit from ensemble methods, for example, by combining multiple outputs into a single output~\citep{huan22la}. There are numerous works on self-correction~\citep{pan23}. For example, self-debugging using self-generated explanations as feedback has been associated with reducing coding errors by LLMs~\citep{chen23}. However, the ability of LLMs to self-correct reasoning by ``reflecting'' on their responses without additional information has also been questioned~\citep{huan23c}. While state-of-the-art systems like GPT-4 improve scores on causal reasoning benchmarks, which can serve as explanations, they also exhibit unpredictable failure modes~\citep{kici23}. 

\subsubsection{Dimensions of Pre-GenAI} \label{sec:preGenAI}
There is also a long list of concepts relevant to our taxonomy that originate from pre-GenAI~\citep{adad18,zini22,dwiv23,schw23,rauk23,sae23a,spei22,minh22}, which we do not elaborate on in detail. For example, XAI methods can also be classified according to what to understand, which includes the system, model-related information such as representations (layers, vectors, embeddings) and outputs, training dynamics, and (impact) of data. This distinction has already been made before GenAI~\citep{schw23}. The most common ways to structure XAI techniques in prior works are, unfortunately, not conceptually clean; for example, a common distinction is between mechanistic and feature attribution-based techniques. But conceptually, feature attribution relates to how the explanation looks (i.e., relevance scores for output). At the same time, mechanistic interpretability aims more at what is being investigated (i.e., neurons and interactions) and how the techniques work (through reverse engineering). As the names of the existing categories are well-established and, thus, easy to comprehend for a reader, we shall also use them in our classification of techniques shown in the next section, but also discuss classification based on our novel dimensions shown in Table \ref{tab:xai_hierarchy}.

\subsection{Classification of Techniques} \label{sec:class}
\begin{figure}[h]
\centering
\includegraphics[width=0.9\textwidth]{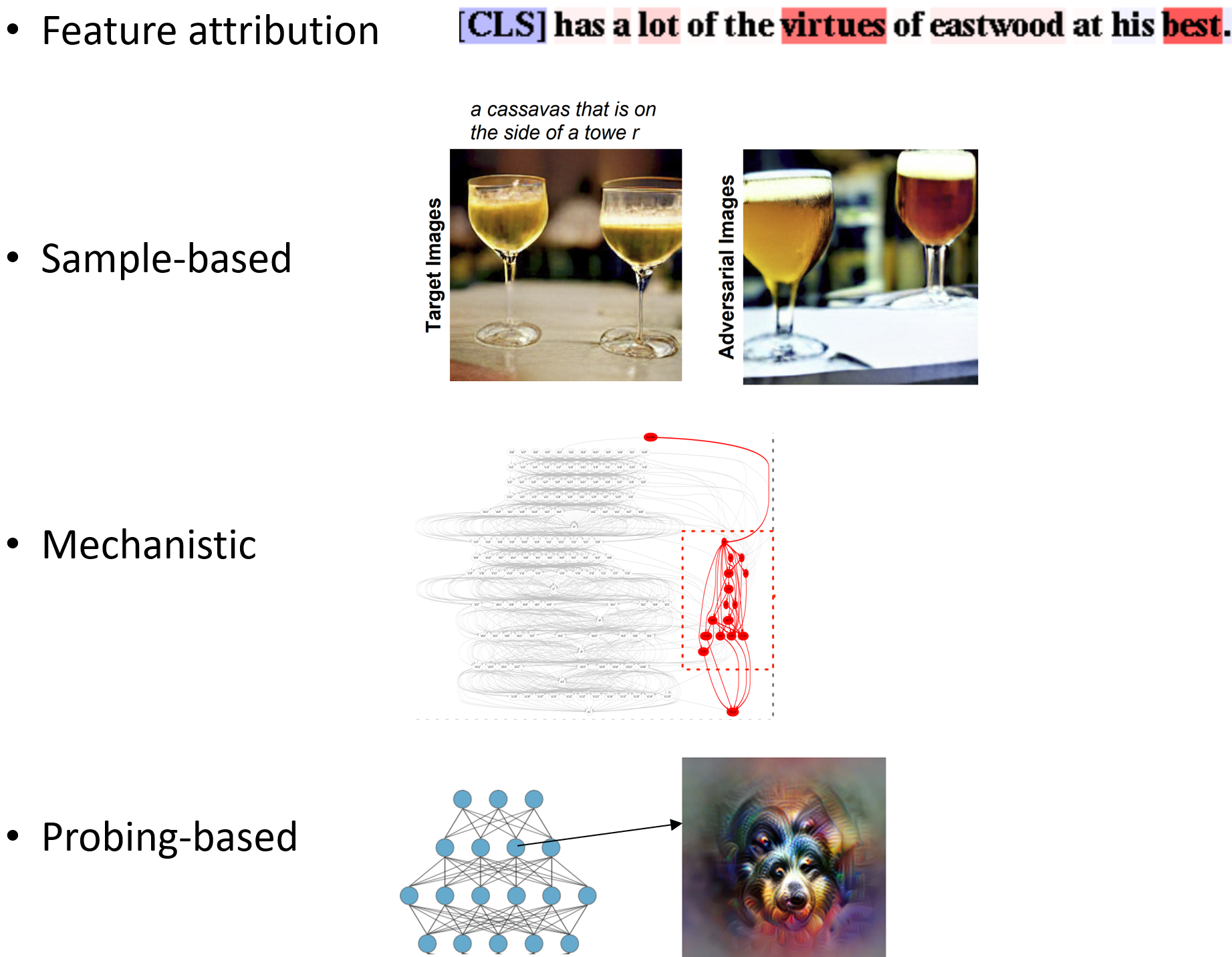}
\caption{Overview of categories of techniques with illustrative examples (Figures are from cited references)}\label{lab:tec}
\end{figure}

We place XAI techniques into four categories commonly found in existing literature (see prior surveys in Section \ref{sec:intro}).
Figure \ref{lab:tec} shows an overview of techniques. Our focus lies on techniques that have been developed particularly for GenAI models or constitute classical techniques that have been adjusted towards GenAI (mostly by addressing computational issues) or could be employed without much change. We also structure existing techniques in terms of our novel dimensions (Section \ref{sec:novDim}).

\subsubsection{Feature Attribution} \label{sec:Attr}
Feature attribution provides a relevance score for each input feature, such as a word or pixel.\\
\noindent\textbf{Perturbation-based techniques} alter inputs partially, for example, by removing or changing features and investigating output (changes). In the context of NLP, alterations such as removing tokens have been investigated \citep{wu2020p} as well as negating and intensifying statements \citep{li2015v}.\\
\noindent\textbf{Gradient-based methods} require a backward pass from outputs to inputs to obtain derivatives. While not all gradient-based techniques are said to work reliably \citep{ade18,ghorb19}, some techniques like Grad-CAM~\citep{selv17} that compute a function of gradients and activations have shown to be valuable on a pixel level in images but also to compute token-level attribution~\citep{mohe21}. Directional gradients have also been used in NLP models~\citep{sikd21,engu23}. Integrated gradients have been used to attribute knowledge to internal neurons~\citep{lun22, dai22} as discussed under ``Neuron activation explanation''. Simple first derivatives concerning embedding dimensions are shown in \citep{li2015v}.\\
\noindent\textbf{Surrogate models} approximate large models using much simpler models - often to understand individual predictions. Classical methods are LIME~\citep{rib16} and SHAP~\citep{lund17}, which have been ported for transformers, that means, for SHAP, see \citep{koka21}. Furthermore, attention flows in NLP models have been shown to be related to SHAP values~\citep{etha21}.
Explain Any Concept (EAC)~\citep{sun24exp} presents an approach for concept explanation, utilizing Segment Anything Model (SAM)~\citep{kir23s} for initial segmentation and introducing a surrogate model to enhance the efficiency of the explanation process. SAM excels at producing object masks from input prompts like partial masks, points, or boxes. It can generate masks for all objects in an image. SAM is trained on a vast dataset that includes 11 million images and 1.1 billion masks. \\
\noindent\textbf{Decomposition-based methods}: Decomposition traditionally refers to attributing relevance from outputs towards inputs or decomposing vectors, but in the context of GenAI, it can also refer to decomposing the reasoning process and attributing outputs to specific reasons. \cite{liu22} aims at explaining the reasoning process for question answers using entailment trees constructed using reinforcement learning. An entailment tree has a hypothesis as its root, reasoning steps as intermediate nodes, and facts as leaves.

Common decomposition techniques compute relevance scores on a layer-by-layer level so that contributions of upper layers emerge as a combination of lower-level contributions. A classic example is Layer-wise relevance propagation (LRP) \citep{monta19}. Decomposition-based methods have also been applied to transformers\citep{ali22}. \cite{ali22} claimed that their adaptation of LRP mitigates shortcomings of gradient methods, which are said to arise due to layernorm and attention layers.
Linear decomposition has also been suggested for local interpretation for transformers \citep{yang23loc}, where a decomposition is considered interpretable if it is orthogonal and linear.
Decompositions are often vector-based~\citep{luo24}. They express a vector (such as a token embedding) in terms of more elementary vectors. For example, \cite{moda23} decomposed (token) vectors and propagated them through the network while maintaining accurate attribution. \cite{zini22} surveyed XAI methods for (word) embeddings. One idea is to express embedding vectors in terms of an (orthogonal) basis of interpretable basis (concept) vectors; another technique employs external knowledge and sparsification of (dense) vectors. \\
\noindent\textbf{Attention-based}: Attention is a key element within neural networks. It provides an importance score for inputs, where inputs are not necessarily inputs to the network but those of a prior layer. For LLMs, attention scores are commonly obtained between all input token pairs for a single attention layer, which can be visualized using a heatmap or bipartite graph~\citep{vig19}. Relevance scores have also been computed by combining attention information and gradients~\citep{bark21}. Attention-based methods have been scrutinized, as they might not identify the most relevant features for predictions~\citep{serr19,jain2019a}. However, the debate has not yet been settled.
\cite{strem22} focus on explaining language models for long texts leveraging sparse attention by developing a masked sampling procedure to identify text blocks contributing to a prediction.\\
Some listed techniques leverage several ideas, for instance, \cite{moda23} can be considered a vector-based and decomposition-based method.

\subsubsection{Sample-based}
Sample-based techniques investigate output changes for different inputs. In contrast to perturbation-based methods that change more selectively individual features to investigate their impact, sample-based techniques focus more on the sample in its entirety to understand the relationship between various inputs and their corresponding outputs rather than attributing the output to specific features within a single input.\\
\noindent\textbf{``Training data influence''} measures the impact of a specific training sample on the model, typically on the output for a particular input to investigate. \cite{gro23st} addressed computational issues to employ influence functions for LLMs. Explainability has been transferred from a large natural language inference dataset to other tasks~\citep{yord21}. \\
\noindent\textbf{Adversarial samples} are input alterations due to small, hard-to-perceive changes for humans that lead to a change in outputs. They are typically discussed in the context of cybersecurity, where an attacker aims to alter model outputs, while a human should not notice the input change. However, also schemes that aim to ``trick'' humans and classifiers alike have been proposed~\citep{sch23dual}. For example, SemAttack~\citep{wang22sem} perturbs embeddings of (tokens) of BERT, while other attacks exchange words~\citep{jin20}. Parts of inputs can also be occluded to understand model behavior better~\citep{sch23conce}.\\
\noindent\textbf{Counterfactual explanations} seek to identify minimal changes to an input so that the output changes from a class $y$ to a specific class $y'$. In contrast to adversarial samples, changes can be noticed by humans. For example, GPT-2 has been fine-tuned to provide counterfactuals based on pairs of original and perturbed input sentences\citep{wu21poly}. Exploring LLM capabilities through counterfactual task variations, \cite{wu23} has shown that LLMs commonly rely on narrow, context-specific procedures that do not transfer well across tasks. \cite{augu22,jean22} use diffusion model guided by classifiers to create counterfactual explanations.
\noindent\textbf{Contrastive explanations} explain why a model predicted $y$ rather than $y'$. Contrastive explanations are said to better disentangle different aspects (such as part of speech, tense, semantics) by analyzing why a model outputs one token instead of another \citep{yin22}.\\

\subsubsection{Probing-based}
Probing-based methods aim at understanding what knowledge an LLM has captured through ``queries'' (probes). A classifier(probe) is commonly trained on a model's activations to distinguish different types of in- and outputs.\\
\noindent\textbf{Knowledge-based}: For example, encoders such as BERT, MiniXX, and T5 that produce vectors can be probed by training a classifier on their outputs that aims to identify the presence of properties of outputs or abilities that emerge from the inputs, such as syntax knowledge~\citep{chen21prob} and semantic knowledge~\citep{tenney19}.
Alternatively to using classifiers, datasets focusing on specific aspects such as grammar~\citep{marv18} can be created. The model's performance on the dataset indicates the ability to capture the property. The design of datasets requires care as regularities might provide an opportunity for shortcut learning \citep{zhon21fac} that foregoes learning the properties in favor of identifying such dataset-specific regularities.
\cite{hern23} learns how to map statements in natural language to fact encodings (in an LLMs representation). In turn, they allow a new way to detect (and explain) when LLMs fail to integrate information from context. The research argues that untruthful texts result from not integrating textual information into specific internal representations.
For example, \cite{liu21pr} investigated the training of RoBERTa over time through probing. They found that local information, such as parts of speech, is acquired before long-distance dependencies such as topics. \cite{goya22} analyzes the learning of text summarization capability of a large language model by obtaining summaries and output probabilities for a fixed set of articles at different points during training. They show n-gram overlap between generated summaries and original articles over time, concluding, for example, that models learn to copy early during training (leading to high overlap), which gets smaller over time (decrease in overlap).\\
\noindent\textbf{Concept-based explanation}: Typically, given a set of concepts, concept-based explanations provide relevance scores of these concepts within inputs \citep{kim18int}. More recently, it has also been proposed to uncover concepts based on what input information is still present at specific layers (or embeddings)~\citep{sch23conce}. While the latter investigates images, high-impact concepts as a source of explanation have also been used for LLMs~\citep{zhao23}. \cite{foote23} allows to interpret a large set of individual neurons by constructing a visualizable graph using the training data and truncation and saliency methods.
However, concept-based methods must be designed carefully, as merely investigating interactions among input variables might be insufficient to show that symbolic concepts are learnt~\citep{li23do}.
\\
\noindent\textbf{Neuron activation explanation}: Individual neurons can also be understood using their activations for inputs. Recently, GPT-4 has been used to generate textual explanations for individual neuron activations of GPT-2~\citep{openai_2023b}. For example, GPT-4 summarizes the text triggering large activations for a neuron.  
\cite{dai22} uncovered ``knowledge neurons'' that store particular facts, i.e., they performed knowledge attribution by setting neuron weights to 0 and increasing it to its original value, while summing up the gradient. If the neuron is relevant for a particular fact, the sum should be large.

\subsubsection{Mechanistic interpretability} \label{sec:mech}
Mechanistic interpretability investigates neurons and their interconnections. It aims at reverse-engineering model components into human-understandable algorithms~\citep{Ola22}. To this end, models can be viewed as graphs~\citep{geig21c}. Circuits, i.e., subgraphs, can be identified that yield certain functionality~\citep{wang2022int}. 
Common approaches fall into three categories~\citep{luo24}, i.e., circuit discovery, causal tracing, and vocabulary lens.
The typical workflow to discover a circuit is often manual and involves~\citep{con24}: (i) Observing a behavior (or task) of a model, creating a dataset to reproduce it, and choosing a metric to measure the extent to which the model performs the task. (ii) Define the scope of interpretation (for example the layers of the model), and (iii) perform experiments to prune connections and components from the model. 
Circuit-based analysis can, thus, also focus on specific architectural elements; for instance, feedforward layers have been assessed and associated with human-understandable concepts~\citep{gev22}. Also, two-layer attention-only has been investigated, leading to conjectures about how in-context learning might work~\citep{ols22}. Recent work has automated finding connections between (abstract) neural network units that constitute a circuit~\citep{con24}. 

(Modern) causal tracing commonly estimates the impact of intermediate activation on their output~\citep{meng22loc}. While causal tracing moves from activation toward outputs, the vocabulary lens focuses on establishing relations to the vocabulary space. For example, \cite{gev22} projects weights and hidden states to the vocabulary space. Individual tokens have also been assessed \citep{ram2022you,katz23i}. \cite{katz23i} create information flow graphs showing (human-readable) tokens based on processed vectors within attention heads and memory values. That is, these vectors are mapped to tokens.

\subsubsection{Structuring based on novel dimensions} \label{sec:novDim}
We also classify existing techniques based on uncovered characteristics, shown in Table \ref{tab:xai_hierarchy}.
Concerning scope, no existing technique explicitly focuses on the entire interaction. However, LLMs providing self-explanations could be used for that purpose. When it comes to explaining one, multiple or all properties of the output, a feature attribution map explains typically at most one attribute of the output, e.g., by highlighting positive and negative words or phrases sentiment might be explained. Sample-based techniques can potentially explain multiple attributes, e.g., if the chosen samples share different characteristics, e.g., for classification if the most similar images show the same object but within a different pose and of different color, it can be concluded that the latter two attributes are irrelevant. Mechanistic interpretability and probing often aim at isolating a single concept, but this is not a must. 

Existing XAI techniques are typically non-dynamics. Explanations from LLMs can be interactive, personalized and are often also sample-dependent.

Figure~\ref{fig:sour} shows a concept matrix linking the foundational source to XAI methods. Feature attribution techniques are typically posthoc methods that leverage the model to generate explanations. However, LLMs could also be prompted to generate feature attribution techniques. Sample-based techniques commonly rely on the training data and the model, for example, to determine which samples most strongly activate a particular neuron. While mechanistic approaches commonly rely on a dataset, it does not have to be data from the training. Probing can be performed naturally through prompts and other forms of providing inputs (and analyzing corresponding outputs).
\begin{table}[h!]
\centering
\begin{tabular}{|c|c|c|c|c||c|c|c|}
\hline
\multicolumn{1}{|c|}{} & \multicolumn{4}{c|}{\textbf{Existing Categories}} & \multicolumn{2}{c|}{\textbf{Selected GenAI Techniques}} \\ \hline

 & \textbf{Feat. Att.} & \textbf{Sample} & \textbf{Mechan.} & \textbf{Prob.} & \textbf{CoT} & \textbf{RAG} \\ \hline
\textbf{Training data} & & x & (x) & & (x) & (x) \\ \hline
\textbf{Model} & x & x & x & x & & \\ \hline
\textbf{Optimization} & & & & x & & \\ \hline
\textbf{Prompts} &(x) & & & x & x & x \\ \hline
\end{tabular}
\caption{Mapping of XAI categories and selected techniques to foundational sources for XAI}\label{fig:sour}
\end{table}

Most existing techniques require white-box access \ref{fig:acc}. It might be possible to gain limited insights using black-box access only, for example, using occlusion (or masking) for feature attribution techniques. It might be investigated if predictions change, but such an approach typically yields coarser and less accurate explanations compared to having access to the output probabilities.

\begin{table}[h!]
\centering
\begin{tabular}{|c|c|c|c|c||c|c|c|}
\hline
\multicolumn{1}{|c|}{} & \multicolumn{4}{c|}{\textbf{Existing Categories}} & \multicolumn{2}{c|}{\textbf{Selected GenAI Techniques}} \\ \hline

 & \textbf{Feat. Att.} & \textbf{Sample} & \textbf{Mechan.} & \textbf{Prob.} & \textbf{Self-explain}& \\ \hline
\multicolumn{5}{|c|}{}\textbf{Required Access} &  &  \\ \hline
\textbf{Black-Box} & (x) &  &  & & x &  \\ \hline
\textbf{White-Box} & x & x & x & x & & \\ \hline
\end{tabular}
\caption{Mapping of XAI categories and selected techniques to required access by XAI techniques}\label{fig:acc}
\end{table}


\section{Research Agenda of XAI for GenAI, Discussion and Conclusions}
As the field of XAI for GenAI is quickly emerging, there are many opportunities for research covering more technical algorithmic avenues onto economic and psychological aspects. Bridging the gap between AI research and other disciplines, such as cognitive science, psychology, and humanities, is likely the way forward for many topics.

\begin{enumerate}
    \item Explaining interactions rather than single input-outputs is particularly interesting for interdisciplinary research such as conducted in the field of information systems and human-computer interaction as it requires both understanding of humans, the model, and the system (See ``Interaction Scope'' in Section~\ref{sec:scope}).
    \item Real-time and interactive explanation based on user queries and feedback could be further explored, necessitating also insights from multiple disciplines (Section~\ref{sec:inter}).  
    \item Multimodal GenXAI: Today, most explanations are of a single modality, i.e., commonly either text or visual. There is a lack of techniques providing explanations using more than one modality (Section~\ref{sec:multimodal}).
    \item Adding XAI to novel directions of GenAI, for instance, GenXAI for video, 3D content generation, and actions (see Table~\ref{tab:aisamp}) is urgently needed.
    \item Deriving novel XAI techniques. In particular, the field of mechanistic interpretability (Section~\ref{sec:mech}) investigating the inner workings of GenAI models is a promising yet challenging frontier.   
    \item Addressing verifiability and hallucinations using AI. Hallucinations are arguably one of the biggest challenges of GenAI. GenXAI can support to mitigate them as seen through techniques such as Chain-of-thought, explaining the reasoning process, but more techniques are needed. 
    \item Porting of pre-GenAI XAI techniques by addressing computational concerns of classical techniques to apply them to large models, as was done for techniques like SHAP (see Section~\ref{sec:Attr}). 
    \item Personalization of explanations: Recognizing the diversity in users' backgrounds, expertise, and needs, future research should focus on personalizing explanations as an important desideratum (see Section~\ref{sec:desi}).
    \item Explanation difficulty has received little attention. A more thorough understanding in the context of GenAI, quantifying the phenomenon and potentially even leading to techniques accounting for such difficulty, is another future avenue (see Section~\ref{sec:expdi}).
    \item Dealing with the complex nature of outputs of GenAI (compared to simple classification in the context of pre-GenAI) remains under-explored. While interactive, user-driven investigation is one avenue in this regard (Section~\ref{sec:inter}) as well as isolating particular facets using mechanistic interpretability techniques focusing on circuits (Section~\ref{sec:mech}), in general, little is known as of today. For example, it is not clear what facets should be explained for text or images (see ``Output Scope'' in Section \ref{sec:scope}) and how to relate actual explanations to high-level objectives such as maximizing plausbability, e.g., ``What attributes should be explained so that an explanation appears plausible?'' (Section~\ref{sec:inter}).
    \item Building GenXAI to target ethical, societal, and regulatory concerns, including contributing to the mitigation of biases and enhancing fairness by identifying and quantifying them through XAI techniques (see second last paragraph in this section).
    \item GenXAI itself raises a number of critical issues with respect to ethical concerns. For example, in legal cases an explanation why someone performed an action can decide over life and death. As such explanation themselves need to carefully assessed with respect to alignment criteria(Section \ref{sec:und}).
\end{enumerate}

As with any other work, this paper employs a particular perspective on the matter of study, foregoing some aspects, and emphasizing others. More concretely, we did not aim to be comprehensive in all regards. Due to the vast array of works on GenAI and XAI, we decided to forego detailed aspects of XAI related to evaluation and usage of explanations. Furthermore, we did not reiterate all existing XAI techniques before GenAI; rather, we focused on novel aspects and methods for GenAI and referred to other surveys for this purpose. Concerning modalities, our core focus was on images and text, driven by the fact that these are (as of now) the two most prevalent modalities. However, others, like video and 3D content generation, are quickly emerging, and audio-to-text (and vice versa) has already been established for some time. We conceptualized existing techniques and discussed some technical aspects, but a more mathematical treatment could be undertaken.
As AI quickly evolves, our work can only be seen as a snapshot in time. Our conceptualization is subject to further evolution, though we believe the uncovered dimensions will pass the test of time, albeit with some enhancements and modifications.
Furthermore,  GenAI (in combination with XAI) touches on many ethical and societal implications, which we partially touched upon in Sections~\ref{sec:cha} and~\ref{sec:und}. However, we refrained from a detailed discussion. We opted to provide a more neutral stance pointing towards concerns rather than being prescriptive and elaborating on actions to be taken to address ethical issues. For example, the lack of access to (commercial) models hinders transparency on the one hand but protects company know-how on the other hand. As such, one might advocate for demanding regulation to increase transparency or double down on allowing companies to be secretive about details regarding their models to protect their intellectual property. From an organizational perspective, more transparency might also support user trust. Thus, the preferred governance of GenAI concerning transparency by companies or from a societal level is debatable~\citep{sch24govco}. We do not take any particular stance in such debates.\\

To conclude, GenXAI is an important area for AI in general. Given the rapid advancement of GenAI and its widespread implications on individuals, society, and economics, more work should be dedicated to this area -- especially given the many research gaps highlighted in our research roadmap elaborating on aspects such as interactivity and scope of explanations. Our work contributes in this direction by providing a thorough motivation for the study of XAI for GenAI, structuring existing knowledge through an enhanced conceptualization of XAI, uncovering novel dimensions, and setting forth a research agenda calling for a joint effort of the research community to address open issues that hopefully contribute to the well-being of all of us.

\section{Acknowledgments}
Two very humble and reputable researchers also contributed to this manuscript but still felt that their contribution did not deserve co-authorship or even be mentioned explicitly in the acknowledgments, which is remarkable in the competitive field of research. Both of whom I want to thank cordially.

\section{Appendix}
\subsection{Research Methodology} \label{sec:resMeth}
Our approach combines aspects of a systematic and a narrative review. However, as explained, XAI for GenAI is rather novel, with a relatively limited number of technical works available. A more qualitative approach was our main focus, for example, a narrative literature review~\citep{king05und}. While we describe our search process below, our goal was not to quantitatively assess peer-reviewed works by showing the evolution and counts of publications over the last years. Currently, works on XAI for GenAI lack a comprehensive conceptualization of important aspects that inform technical works, such as challenges and desiderata for explanations for GenAI. Therefore, the importance and challenges we identified are a synthesis of existing works and a creative combination taking essential properties of GenAI and prior XAI desiderata into account. Our taxonomy is based on (i) a meta-survey of (pre-)GenAI taxonomies and surveys, as many concepts and characteristics are still relevant for GenAI, as well as (ii) novel works contributing towards XAI for GenAI. We searched for several terms on Google Scholar between the 15th of February 2024 and the 15th of March 2024 using either ``survey'' or ``review'' in  ``Survey explainability'', ``Survey XAI for generative AI'', ``Survey explainability for large language models'', ``explainability large language models'', ``explainability transformer'', ``explainability diffusion model''.  This was followed by forward and backward search~\citep{web02}.
We filtered results based on title, abstract, and full-text as follows: We preferred peer-reviewed works but also commonly included articles from arxiv.org after performing our quality assessment as reviewers. We were stricter in inclusion with older works (on pre-GenAI) and more open to works on XAI for GenAI containing novel aspects as there are fewer works, and we deem it important to include ideas that are still in the making. Our taxonomy development process followed that of \cite{nick13} that is we alternated between looking at concepts (mostly synthesized in earlier surveys) and empirical data (primary research papers on XAI for GenAI) to derive our dimensions iteratively and, ultimately, our taxonomy.

\bibliography{sn-bibliography}
\end{document}